\documentclass[letterpaper,10pt,conference]{IEEEtran}

\IEEEoverridecommandlockouts

\usepackage{combelow}

\usepackage{glossaries}
\newacronym{fmcw}{FMCW}{Frequency-Modulated Continuous-Wave}
\newacronym{nn}{NN}{neural network}
\newacronym{fft}{FFT}{Fast Fourier Transform}
\newacronym{cnn}{CNN}{Convolutional Neural Network}

\usepackage{cite}

\usepackage{graphicx}

\usepackage{paralist}

\usepackage[labelformat=simple]{subcaption}

\usepackage[hidelinks]{hyperref}
\newcommand\rurl[1]{%
  \texttt{\href{http://#1}{\nolinkurl{#1}}}%
}

\usepackage{cleveref}
\crefname{table}{Tab.}{Tabs.}
\crefname{figure}{Fig.}{Figs.}
\crefname{section}{Sec.}{Secs.}
\crefname{equation}{Eq.}{Eqs.}

\usepackage[binary-units]{siunitx}
\usepackage{pgfplotstable}
\usepackage{tikz}

\usepackage{amssymb}
\usepackage{mathabx}

\usepackage{xcolor}

\usepackage{copyright}

\begin{document}

\title{
\Large \bf Contrastive Learning for Unsupervised Radar Place Recognition
}
\author{
Matthew Gadd, Daniele De Martini, and Paul Newman\\
Mobile Robotics Group (MRG), University of Oxford.\\\texttt{\{mattgadd,daniele,pnewman\}@robots.ox.ac.uk}
}
\maketitle

\copyrightnotice

\begin{abstract}
We learn, in an unsupervised way, an embedding from sequences of radar images that is suitable for solving the place recognition problem with complex radar data.
Our method is based on invariant instance feature learning but is tailored for the task of re-localisation by exploiting for data augmentation the temporal successivity of data as collected by a mobile platform moving through the scene smoothly.
We experiment across two prominent urban radar datasets totalling over \SI{400}{\kilo\metre} of driving and show that we achieve a new radar place recognition state-of-the-art.
Specifically, the proposed system proves correct for \SI{98.38}{\percent} of the queries that it is presented with over a challenging re-localisation sequence, using only the single nearest neighbour in the learned metric space.
We also find that our learned model shows better understanding of out-of-lane loop closures at arbitrary orientation than non-learned radar scan descriptors.
\end{abstract}

\section{Introduction}%
\label{sec:introduction}

For autonomous vehicles to drive safely at speed, in inclement weather, or in wide-open spaces, very robust sensing is required.
Thus, the interest in scanning \gls{fmcw} radar for place recognition and localisation.
However, all of the approaches presented thus far which are learned from data are also supervised.
As more datasets featuring this modality become available, unsupervised techniques will allow us to learn from copious unlabelled measurements.
Nevertheless, there are interesting sensor characteristics that must be overcome in order to fully exploit this technology for mapping and localisation tasks.
These include artefacts and speckle noise, sensor configuration, perceptual ambiguity and aliasing, and road layout and driving scenarios.

\begin{figure}
\centering
\includegraphics[width=0.75\columnwidth]{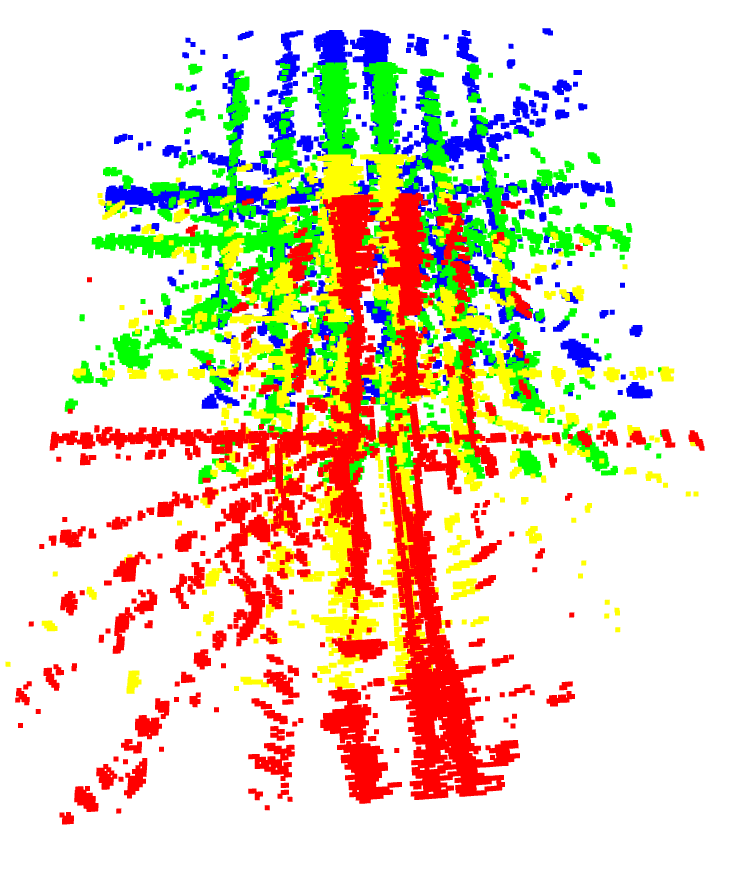}
\vspace{-.3cm}
\caption{
Rendering of the radar scan content with which our system learns how to perform place recognition.
Our approach is based on unsupervised feature learning which augments mini-batch instances, forcing augmentations and instances to be close in a metric space and instances to be far away from each other.
In applying this to the radar domain, we improve this feature learning approach by making sure to include in the batch instances which are related by movement but are nevertheless farflung, shown red and blue.
Augmentations themselves are not -- as is typical -- synthetically generated from each scan instance, but are generated from nearby scans along the radar video, shown yellow and green.
Here, radar scans are converted to pointclouds by thresholding on a power floor for visualisation purposes -- they are fed to our network as imagery, however.
Generated with Open3D~\cite{Zhou2018}.
}
\label{fig:painted}
\vspace{-.6cm}
\end{figure}

\cref{fig:painted} illustrates a sequence of temporally successive radar scans used to train our system.

As in previous work~\cite{saftescu2020kidnapped,gadd2021unsupervised}, we do not pursue fine-grained localisation with pose refinement but are interested in learning robust \emph{representations} of radar imagery, which are quick to compute and compare and which will in any case bolster the performance of downstream pose estimation systems.
In our work, we exploit the fact that the sensory records collected by our vehicle are strongly sequential within an otherwise unsupervised learning framework.
Specifically, we leverage temporally successive complex radar data as ``augmentations'' of instances in mini-batches in order to learn robust and invariant radar scan embeddings.
We exploit commonality in the sensing horizons in every direction around the robot of scans captured shortly after one another in radar videos
This starts from significant overlap (yellow on red) and degrades to minimal overlap (blue on red).
The primary effect of this strategy is that our learned network is presented with more \emph{real} data than typical contrastive learning approaches, leveraging the ``noises'' in radar measurements (c.f. artefacts and speckle noise, above).
Indeed, maximally overlapping scans are a rich source of training data for neural networks, as opposed to totally synthesised contrastive content.
Minimally overlapping scans are useful in learning a metric space -- we should ensure that their embeddings are well separated (c.f. perceptual ambiguity and aliasing, above).
This work is an extension of our initial investigation in~\cite{gadd2021unsupervised}, where we additionally boost performance by measuring place-to-place distance as the distributional similarity of stochastic embeddings generated by unfreezing dropout layers during inference.

Our \emph{principal contributions} are as follows:
\begin{enumerate}
\item Mini-batch strategies which exploit the temporal succession of radar scans gathered from the scanning sensor, 
\item Augmentation operations for learning invariance to rotation, 
\item Boosting performance by using active dropout layers during inference and a distributional similarity metric which accounts for uncertainty due to the stochasticity of the radar measurement process as well as learned weights.
\end{enumerate}

Incidentally, we perform the first tests of a place recognition system across multiple radar datasets (c.f. sensor configuration, above), and comment on the suitability of extant metrics and the presentation of a more nuanced interpretation decomposed into common driving scenarios (c.f. road layout and driving scenarios, above).

\section{Related Work}%
\label{sec:related}

Radar place recognition and localisation has received some interest, albeit not as much as the related task of motion estimation.
Kim \textit{et al}~\cite{kim2020mulran} show that radar can outperform LiDAR in place recognition using a non-learned rotationally-invariant ring-key descriptor.
S{\u{a}}ftescu \textit{et al}~\cite{saftescu2020kidnapped} present the first supervised deep approach, adapting \acrshortpl{cnn} for equivariance and invariance to azimuthal perturbations.
Barnes and Posner~\cite{barnes2020under} learn keypoints which are useful for motion estimation and localisation simultaneously.
Gadd \textit{et al}~\cite{gadd2020lookaroundyou} leverage rotationally-invariant representations in a sequence-based localisation system.
De Martini \textit{et al}~\cite{demartini2020kradar} fine-tune the results from the embedding space with pointcloud registration techniques.
Wang \textit{et al} learn metric localisation directly along with self-attention~\cite{wang2021radarloc}.

Unsupervised visual place recognition and image retrieval is an emerging area of study.
Merrill and Huang~\cite{merrill2018lightweight} use an autoencoder tasked with decoding a handcrafted HOG descriptor, as a geometric prior.
Caron \textit{et al}~\cite{caron2018deep} iteratively assign pseudo-labels to images by clustering deep features periodically while training.
Zaffar \textit{et al}~\cite{zaffar2020cohog} propose image areas of focus based on entropy measures and describe the image by regional descriptors which are convolutionally matched for viewpoint-invariance.
Bonin-Font and Burguera~\cite{bonin2021nethaloc} present an approach in which each image in the training dataset has loop closured candidates synthesised from automatic random rotation over the image plane, scaling and shifting.

Contrastive learning in place recognition is  increasingly studied.
In~\cite{leyva2019place}, Leyva-Vallina \textit{et al} a contrastive loss function and inputs of similar or dissimilar image pairs are used while in~\cite{leyva2021generalized} generalised contrastive loss function that relies on image similarity as a continuous measure is used to train a siamese network.
Radenovi\'{c} \textit{et al}~\cite{radenovic2018fine} propose a novel trainable pooling layer that generalises max and average pooling and show that it boosts retrieval performance.
Sermanet \textit{et al}~\cite{sermanet2018time} obtain anchor and positive images taken from simultaneous viewpoints are encouraged to be close in the embedding space, while distant from negative images taken from a different time in the same sequence.

In addition, three prominent urban datasets have recently been released, focused on this modality~\cite{barnes2020rrcd,kim2020mulran,sheeny2020radiate}.
The \textit{RADIATE}~\cite{sheeny2020radiate} dataset is more suited to object detection and tracking.
\textit{MulRan}~\cite{kim2020mulran} provides ground truth poses, but does not repeat the same routes very many times.
The \textit{Oxford Radar RobotCar Dataset}~\cite{barnes2020rrcd} is ideal for developing and testing place recognition and localisation systems in that it revisits the same route on many occasions.

\section{Method}%
\label{sec:method}

Our method is illustrated in~\cref{fig:system}, where in the discussion below instances (referring to Cartesian radar scans) and embeddings (referring to feature vectors) are denoted $\mathbf{x} \in \mathbb{R}^{W{\times}W}$ (obtained from polar scans of size $A{\times}B$) and $\mathbf{f} \in \mathbb{R}^{d}$, respectively.
\cref{sec:featlearn} describes the baseline system we start with while \cref{sec:motaug,sec:distsim} describes the proposed sampling/augmentation strategies and similarity measurement.

\subsection{Invariant Instance Feature Learning}
\label{sec:featlearn}

\begin{figure}[!h]
\centering
\includegraphics[width=\columnwidth]{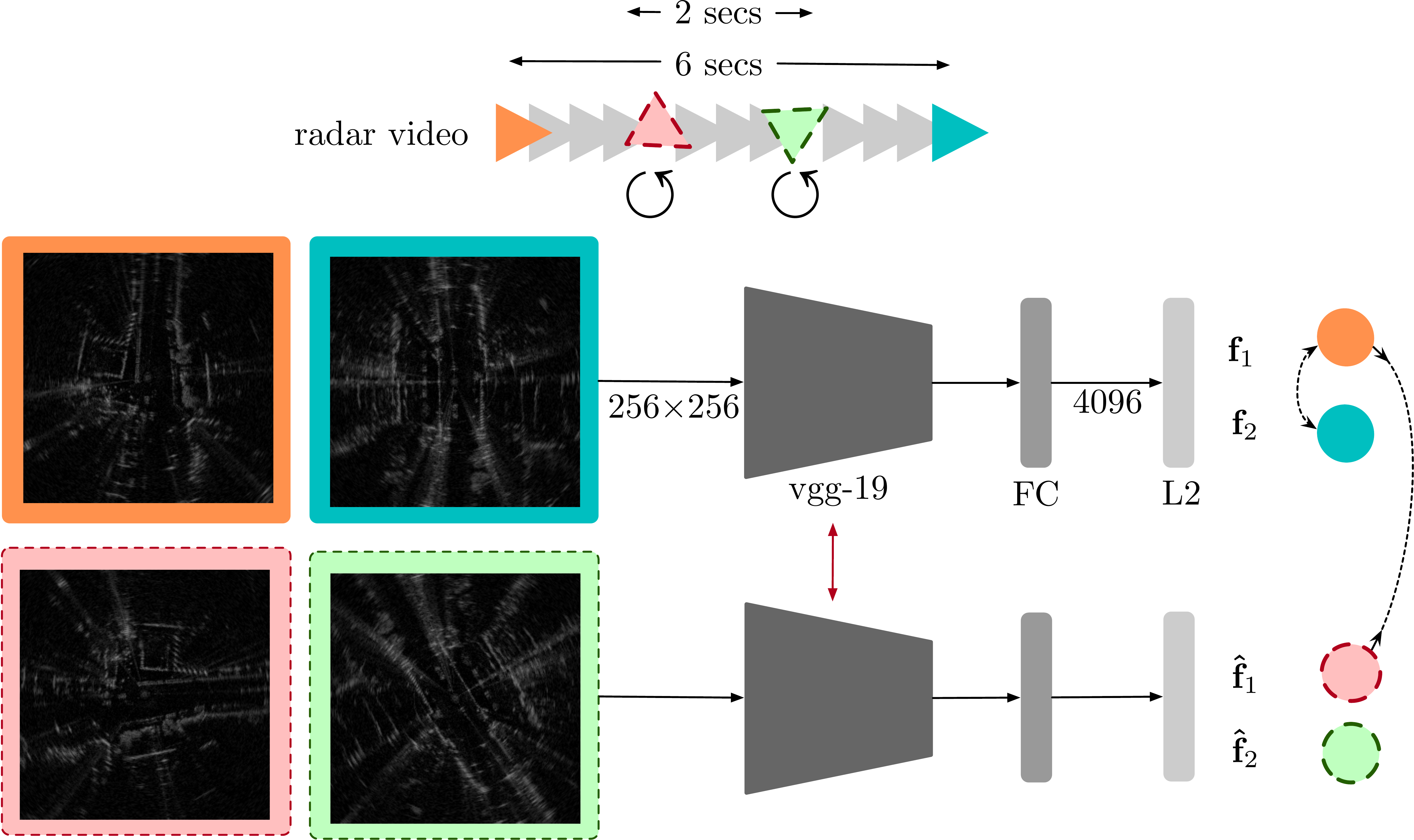}
\caption{
Overview of our network architecture and the training procedure more generally -- described in~\cref{sec:method}.
}
\label{fig:system}
\vspace{-.5cm}
\end{figure}

We adapt the unsupervised embedding learning approach of Ye \textit{et al}~\cite{ye2019unsupervised}\footnote{with loss implemented as per \rurl{github.com/mangye16/Unsupervised_Embedding_Learning}}, to ensure that features of distinct instances are separated (\cref{fig:system} right, arrow line between orange and blue) while features of an augmented instance are invariant (\cref{fig:system} right, back-arrow line between orange and red).
To achieve this, a mini-batch is constructed using the instances $\{\mathbf{x}_1, \mathbf{x}_2, \ldots, \mathbf{x}_m\}$ and their augmentations, $\{\hat{\mathbf{x}}_1, \hat{\mathbf{x}}_2, \ldots, \hat{\mathbf{x}}_m\}$.
The probability of the augmentation, $\hat{\mathbf{x}}_i$, of an instance, $\mathbf{x}_i$, being recognised as that instance is

\begin{equation}
P(i \vert \hat{\mathbf{x}}_i) = \frac{\mathrm{exp}(\frac{\mathbf{f}_i^T\hat{\mathbf{f}_i}}{\tau})}{\sum_{k=1}^m\mathrm{exp}(\frac{\mathbf{f}_k^T\hat{\mathbf{f}_i}}{\tau})}
\end{equation}

The probability of another instance, $\mathbf{x}_j$, being recognised as the instance $\mathbf{x}_i$, is given by 

\begin{equation}
P(i \vert \mathbf{x}_j) = \frac{\mathrm{exp}(\frac{\mathbf{f}_i^T\mathbf{f}_j}{\tau})}{\sum_{k=1}^m\mathrm{exp}(\frac{\mathbf{f}_k^T\mathbf{f}_j}{\tau})},~j \neq i
\end{equation}

We need to enforce that the instance augmentation be classified as the instance and that no other instance is likewise classified. 
The joint probability of this is given by

\begin{equation}
P_i = P(i \vert \hat{\mathbf{x}}_i) \prod _{j \neq i}(1-P(i \vert \mathbf{x}_j))
\end{equation}

Therefore, over the entire batch, we seek to minimise 

\begin{equation}
\label{eqn:objective}
J = -\sum_i \mathrm{log}P(i \vert \hat{\mathbf{x}}_i) - \sum_i \sum_{j \neq i}\mathrm{log}(1-P(i \vert \mathbf{x}_j))
\end{equation}

\subsection{Motion-enabled data augmentation}
\label{sec:motaug}

The baseline approach in~\cref{sec:featlearn} is typically applied using synthetically generated instances of images.
We can do better than this in the place recognition task, by exploiting the strong signal from the relationship between scans captured close in time.

We retrieve an instance randomly by using a shuffled sampler to draw a sample from the trajectory where each $\mathbf{x}_i$ is equally likely to be drawn, as is standard in deep learning procedures (\cref{fig:system}, orange).
We then retrieve a scan several frames ahead of this instance, $\hat{\mathbf{x}}_i$, to within a maximum time difference of $d_{min} = $~\SI{2}{\sec} (\cref{fig:system}, red).

\begin{equation}
\hat{\mathbf{x}}_i = \mathbf{x}_{i+k},~k \in [0, d_{min}]
\end{equation}

This takes the place of the ``augmentation'' which in other applications would be identical to the instance $\mathbf{x}_i$.
This ensures that real data is presented to the network, rather than synthetic augmentations which would anyway be prohibitively difficult to simulate for radar sensor artefacts.
This also presents the network with some translational invariance, along the route driven.
This ``augmentation'' is further augmented by rotating it  (\cref{fig:system}, red, $\circlearrowleft$).

\begin{equation}
\hat{\mathbf{x}}_i = \mathrm{R}(\mathbf{x}_{i+k}, r),~k \in [0, d_{min}],~r \in [0,A]
\end{equation}

where $A$ is the number of azimuths in the polar grid and $\mathrm{R}$ denotes a cylindrical shift of the azimuthal returns.
This helps us achieve rotational invariance.
True scans at similar attitudes are of course available in the dataset but would only be usable in a supervised learning framework, and so we settle for synthetic data augmentation at this point.
Finally, other batch instances $\mathbf{x}_j,~j \neq i$ (\cref{fig:system}, blue) and their augmentations $\hat{\mathbf{x}}_j$ (\cref{fig:system}, green, $\circlearrowleft$) are retrieved at least as distant (in time) as $d_{min} = $~\SI{2}{\sec} and at most as distant (in time) as $d_{max} = $~\SI{6}{\sec} from the original instance.

\begin{equation}
\begin{split}
\hat{\mathbf{x}}_j = \mathbf{x}_{j+k},~k \in [d_{min}, d_{max}]\\
\hat{\mathbf{x}}_j = \mathrm{R}(\mathbf{x}_{j+k}, r),~k \in [d_{min}, d_{max}],~r \in [0,A]
\end{split}
\end{equation}

This mitigates the (already quite small but non-zero) chance that batches may simultaneously hold instances which are in fact nearby, although randomly sampled.
We also expect that because these instances are somewhere nearby each other with some data overlap that the constraint of~\cref{eqn:objective} enforces a better metric space where they are well separated and are nevertheless distinct places.
Sampling based on attenuated global positioning is also possible but we present this strategy in terms of sampling based on elapsed scans in the radar video as it is truer to the unsupervised nature of the base system that we adapt to this task.
It should be noted that the base system, Ye \textit{et al}, is prone to instances in the batch being in fact of the same class, analogously to the vehicle idling in the same position in our system.
The second sample at up to $d_{max}$ reduces the likelihood of this as compared to using $d_{min}$ alone.
\cref{sec:results} proves the effectiveness of each sampling and data augmentation step, which are designed to counteract stationary data.

\subsection{Measuring place similarity}
\label{sec:distsim}

\begin{figure*}[!h]
\centering
\begin{subfigure}{0.42\columnwidth}
\includegraphics[width=\textwidth]{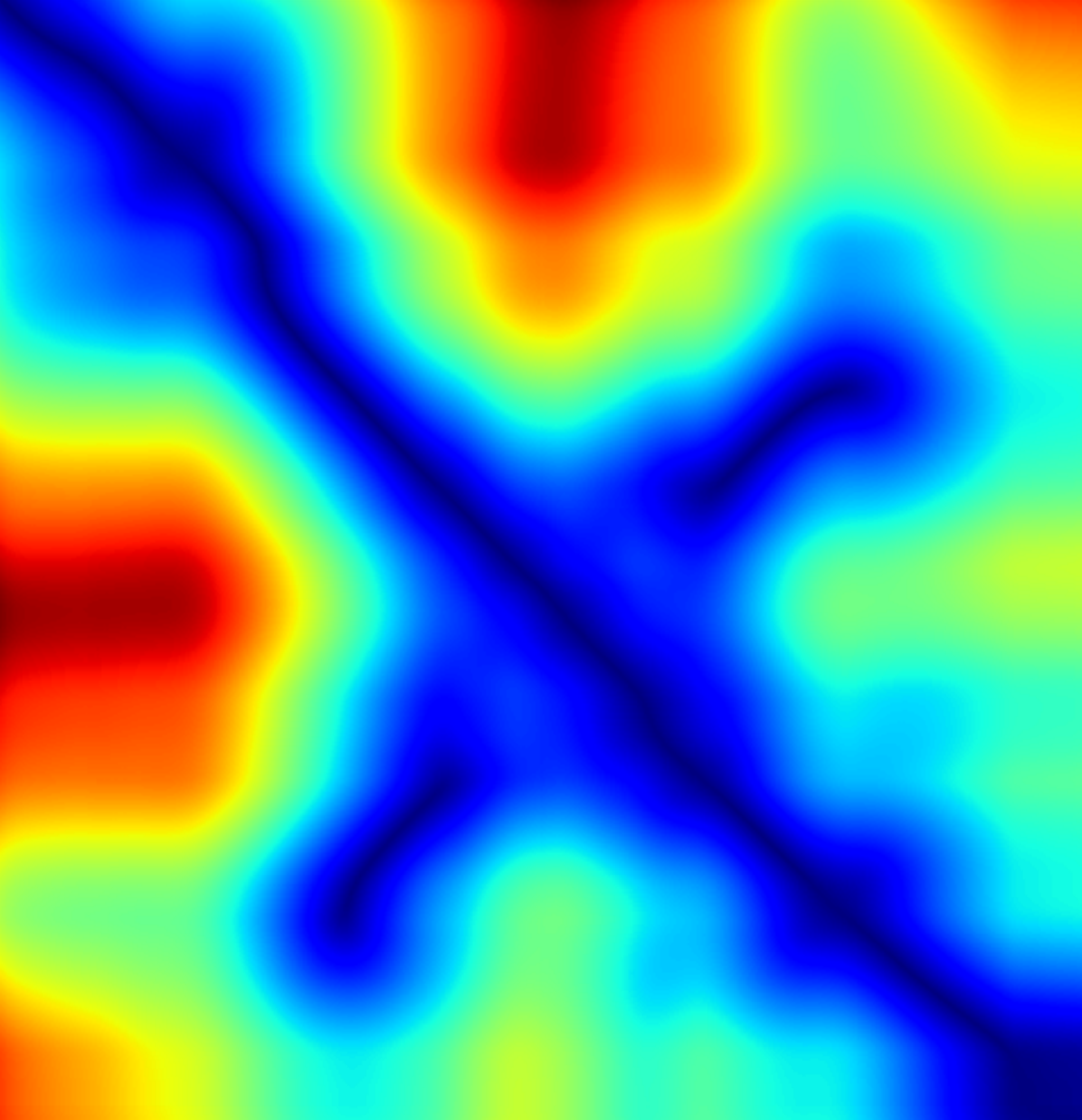}
\caption{}
\label{fig:gt}
\end{subfigure}
\begin{subfigure}{0.42\columnwidth}
\includegraphics[width=\textwidth]{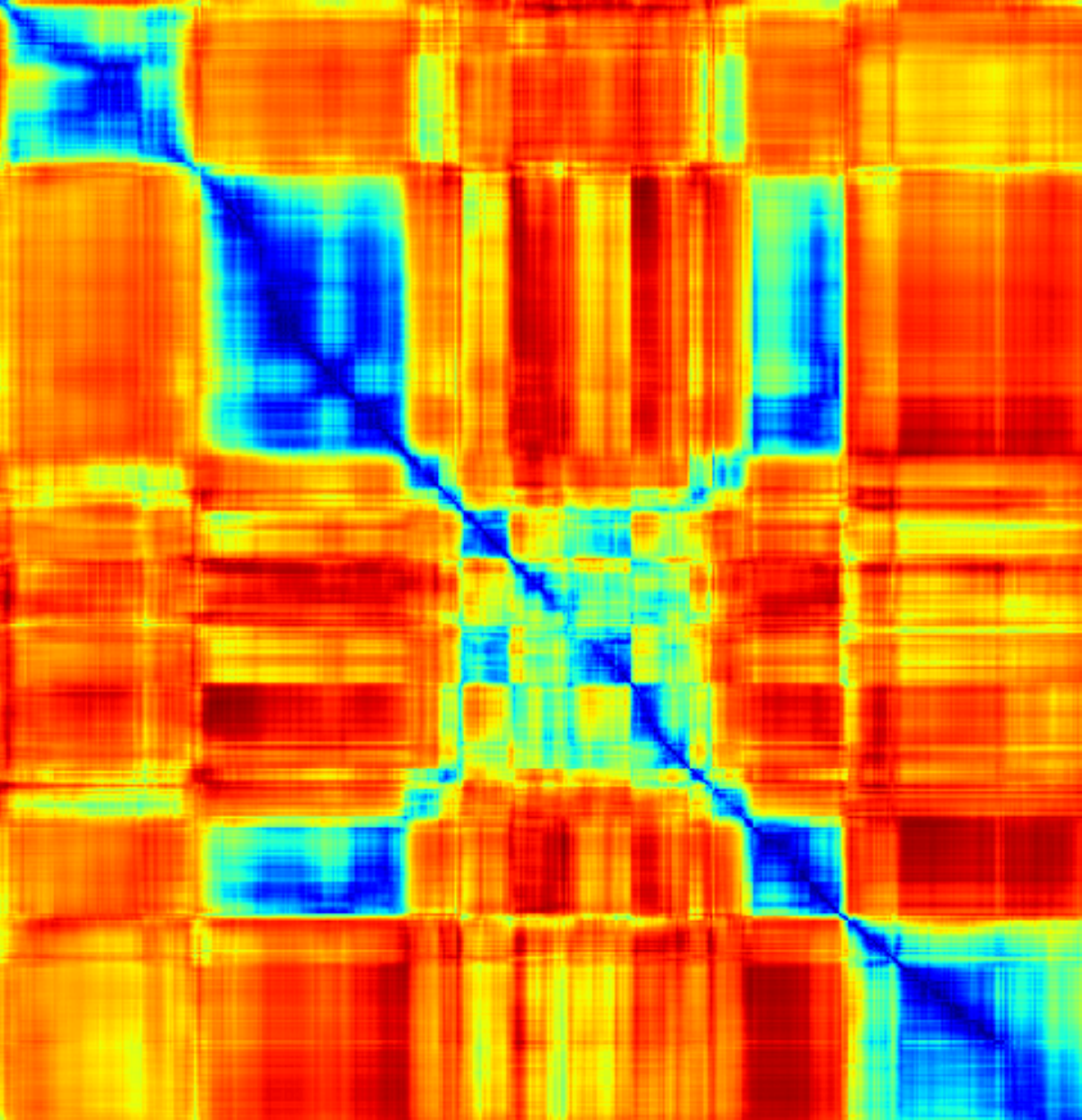}
\caption{}
\label{fig:vTR2_dist}
\end{subfigure}
\begin{subfigure}{0.42\columnwidth}
\includegraphics[width=\textwidth]{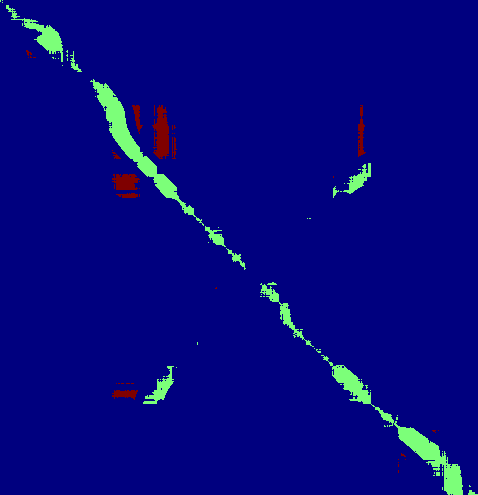}
\caption{}
\label{fig:bsnn}
\end{subfigure}
\begin{subfigure}{0.42\columnwidth}
\includegraphics[width=\textwidth]{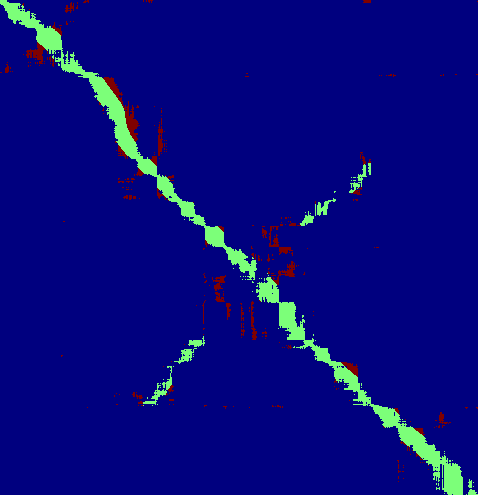}
\caption{}
\label{fig:rnnn}
\end{subfigure}
\caption{
\subref{fig:gt} Ground truth matrix,
\subref{fig:vTR2_dist} Distance matrix for embeddings learned by our top-performing model, \texttt{vTR2}, 
\subref{fig:bsnn} match matrix for \texttt{Recall@P=80\%},
\subref{fig:rnnn} match matrix for \texttt{Recall@N=25}.
In \subref{fig:bsnn} and \subref{fig:rnnn}, true positives are shown green while false positives are shown red.
}
\label{fig:mats}
\vspace{-.5cm}
\end{figure*}

In performing place recognition by neighbourhood searches in the learned metric space -- i.e. in constructing difference matrices such as those shown in~\cref{fig:mats}, for a fixed embedding function we typically use Euclidean or cosine distance metrics.
In this work, in order to account for uncertainty in the model weights, stochasticity of the radar measurement process, and perceptual ambiguity of self-similar scenes we produce a family of embeddings for each map and query frame during inference.
This family of embeddings is available by activating dropout layers during inference.
We then use as a distance metric the Kullback-Liebler divergence between two normal distributions which are located at the mean and scaled by the variance of each embedding family.
While we only explore Kullback-Liebler, other divergence measures -- such as Mahalanobis -- are also appropriate here.

\subsection{Training details and parameters}

Polar radar scans of with $A = 400$ azimuths and $B = 3768$ bins of size \SI{4.38}{\centi\metre} are converted to Cartesian scans with side-length $W = 256$ and bin size \SI{0.5}{\metre}.
Feature vectors at the output of our networks are length $d = 4096$.
We train each variant for \num{10} epochs, using a learning rate of $3\mathrm{e}{-4}$, batch size of \num{12}.
We collect $T = 24$ samples for distributional parameters during forward passes with active dropout.
For comparing against the non-learned descriptor of Kim \textit{et al}~\cite{kim2020mulran}, we use the reported ring-key parameters which downsample the radar scan to $120{\times}40$ before averaging over azimuths to form a $40{\times}1$ feature vector.

\section{Experimental Setup}%
\label{sec:experiments}

\subsection{Datasets}

We train and test our method with various urban radar data.

\subsubsection{Oxford}

These data are collected in the \textit{Oxford Radar RobotCar Dataset}~\cite{barnes2020rrcd}.
Train-test splits are as per~\cite{saftescu2020kidnapped}, testing on hidden data featuring backwards traversals, vegetation, and ambiguous urban canyons\footnote{\texttt{2019-01-18-14-46-59-radar-oxford-10k} for mapping and \texttt{2019-01-18-15-20-12-radar-oxford-10k} for localisation}.
We do, however, train on most data available from~\cite{barnes2020under} (barring anything from the test split) -- totalling \num{30} forays, or about \SI{280}{\kilo\metre} of driving.
This is a similar quantity of training data as reported in~\cite{barnes2020under}.

\pgfplotsset{
select row/.style={
x filter/.code={\ifnum\coordindex=#1\else\def\pgfmathresult{}\fi}
}
}

\pgfplotstableread[col sep=comma]{
metric,vT,vR,vTR,vTR2
95,7.081667618466944,0.5477389543608965,12.400706089962094,16.349099215947515
98,4.262499350998065,0.5477389543608965,9.041586625777262,13.208036965821046
99,4.262499350998065,0.4880328124163437,7.533357561873561,11.728363013282133
}\prOxfordData

\pgfplotstableread[col sep=comma]{
metric,vT,vR,vTR,vTR2
1,97.37373737373738,73.13131313131314,96.16161616161617,98.38383838383838
2,98.7878787878788,83.03030303030303,96.76767676767678,99.19191919191918
3,99.19191919191918,87.67676767676768,97.77777777777777,99.19191919191918
}\rcOxfordData

\begin{figure*}[!h]
\centering

\begin{subfigure}{0.48\textwidth}
\begin{tikzpicture}
\begin{axis}[
ybar,
bar width=.5cm,
width=\columnwidth,
height=.375\textwidth,
enlarge x limits=0.25,
legend style={at={(0.7,1.4)}, anchor=north,legend columns=-1},
symbolic x coords = {95,98,99},
xtick=data,
xlabel={$P$ (\%)},
ylabel=\texttt{Recall@P} (\%),
grid=major,
xmajorgrids=false
]
\addplot table[x=metric,y=vR]{\prOxfordData};
\addplot table[x=metric,y=vT]{\prOxfordData};
\addplot table[x=metric,y=vTR]{\prOxfordData};
\addplot table[x=metric,y=vTR2]{\prOxfordData};
\legend{\texttt{vR},\texttt{vT},\texttt{vTR},\texttt{vTR2}}
\end{axis}
\end{tikzpicture}
\caption{
\texttt{Recall@P} metrics for the \textit{Oxford} dataset.
}
\label{fig:pr-oxf}
\end{subfigure}
\begin{subfigure}{0.48\textwidth}
\begin{tikzpicture}
\begin{axis}[
ybar,
bar width=.5cm,
width=\columnwidth,
height=.375\textwidth,
enlarge x limits=0.25,
legend style={at={(0.7,1.4)}, anchor=north,legend columns=-1},
symbolic x coords = {1,2,3},
xtick=data,
xlabel={$N$ - Database candidates},
ylabel=\texttt{Recall@N} (\%),
grid=major,
xmajorgrids=false
]
\addplot table[x=metric,y=vR]{\rcOxfordData};
\addplot table[x=metric,y=vT]{\rcOxfordData};
\addplot table[x=metric,y=vTR]{\rcOxfordData};
\addplot table[x=metric,y=vTR2]{\rcOxfordData};
\legend{\texttt{vR},\texttt{vT},\texttt{vTR},\texttt{vTR2}}
\end{axis}
\end{tikzpicture}
\caption{
\texttt{Recall@N} metrics for the \textit{Oxford} dataset.
}
\label{fig:rc-oxf}
\end{subfigure}

\caption{
Evaluation of \texttt{Recall@P} and \texttt{Recall@N} performance for batch strategies over the \textit{Oxford Radar RobotCar Dataset}.
}
\label{fig:batchStrats}
\vspace{-.6cm}
\end{figure*}
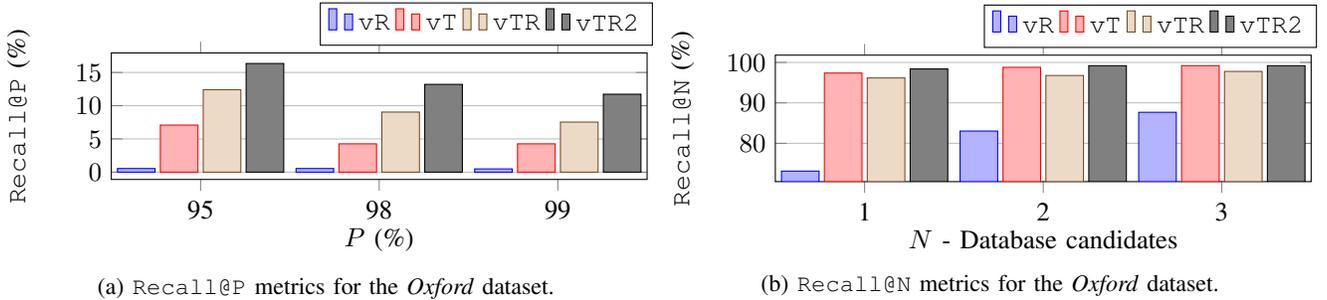

\subsubsection{MulRan}

These data are collected in the \textit{MulRan} dataset~\cite{kim2020mulran}.
We train using all but one of these sequences\footnote{\texttt{DCC}, \texttt{KAIST}, \texttt{Sejong}}.
The test sequence\footnote{\texttt{Riverside}} is chosen as it features revisits in the opposite direction as well as lane-level differences.
The data used totals approximately \SI{124}{\kilo\metre} of driving.

\subsection{Benchmarking}

In addition to the distributional similarity (denoted \texttt{KL}), our ablation study examines the performance gain available from the motion-enabled strategies described in~\cref{sec:motaug} above, namely:
\begin{itemize}
\item \texttt{vR} or ``spin augmenting'': where augmentations are produced from instances themselves and synthetic rotations,
\item \texttt{vT} or ``video sampling'': where augmentations are taken from temporally proximal frames in the radar sequence, but not synthetically rotated,
\item \texttt{vTR} or ``video sampling and spin augmenting'': temporally proximal frames which are also synthetically rotated, and finally
\item \texttt{vTR2}: or ``video sampling and spin augmenting with batch pairing'', where other instances in the batch are ensured to be true negatives.
\end{itemize}
Note the baseline of~\cite{ye2019unsupervised}, entails augmenting \emph{only} the instance itself, closest in spirit to \texttt{vR}.

We also compare our method via direct experimentation to a non-learned radar scan descriptor using the method of Kim \textit{et al}~\cite{kim2020mulran} as it is readily implementable and not learned from data.
To embed the radar scan, this method resizes it and averages within range-bins and over azimuths to form a feature vector.
We refer to these results as \texttt{SC}.
Finally, we also discuss performance figures in comparison to results reported in S{\u{a}}ftescu \textit{et al}~\cite{saftescu2020kidnapped} and Barnes and Posner~\cite{barnes2020under}.

For completeness, we note that we use either \texttt{VGG-19}~\cite{simonyan2014very} or \texttt{ResNet-152}~\cite{he2016deep} as front-end feature extractors\footnote{implementations based on \rurl{github.com/pytorch/vision}}.

\subsection{Performance assessment}
For performance assessment, we employ both \texttt{Recall@P} -- where predicted positives are within in a varying embedding distance threshold of the query -- and \texttt{Recall@N} -- where a query is considered correctly localised if \emph{at least one} database candidate is close in space as measured by ground truth (in our case, GPS/INS).
Note that the latter is a more lenient measure of performance and that we employ no pose refinement to boost results, as in~\cite{kim2020mulran,demartini2020kradar}.
We specifically impose -- for both \textit{Oxford} and \textit{MulRan} datasets -- for the first metric (e.g. \cref{fig:pr-oxf,fig:Datasets}) boundaries of \SI{25}{\metre} and \SI{50}{\metre}, inside of which predicted matches are considered true positives, outside of which false positives, as per~\cite{saftescu2020kidnapped}.
The second metric (e.g. \cref{fig:rc-oxf,fig:netArch,fig:Decomp,fig:Uncertain}) supports only a single boundary, which we set to \SI{25}{\metre}, the more difficult of the two.
For precision-recall results, we also present aggregate metrics, including F-scores ($F_{0.5}$ $F_1$, and $F_2$) and area-under-curve (\texttt{AUC}).

\subsection{Loop closure event types}
\label{sec:decom}

\begin{figure}[!h]
\centering
\begin{subfigure}{0.36\columnwidth}
\includegraphics[width=\textwidth]{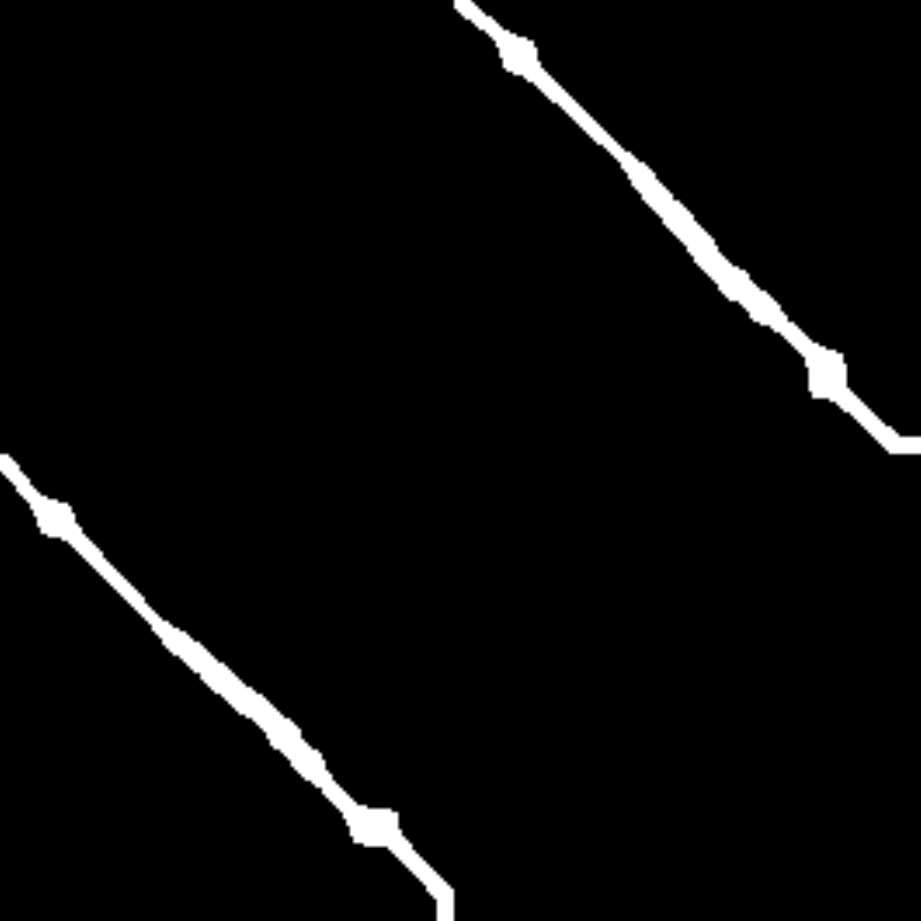}
\caption{}
\label{fig:R2}
\end{subfigure}
\begin{subfigure}{0.36\columnwidth}
\includegraphics[width=\textwidth]{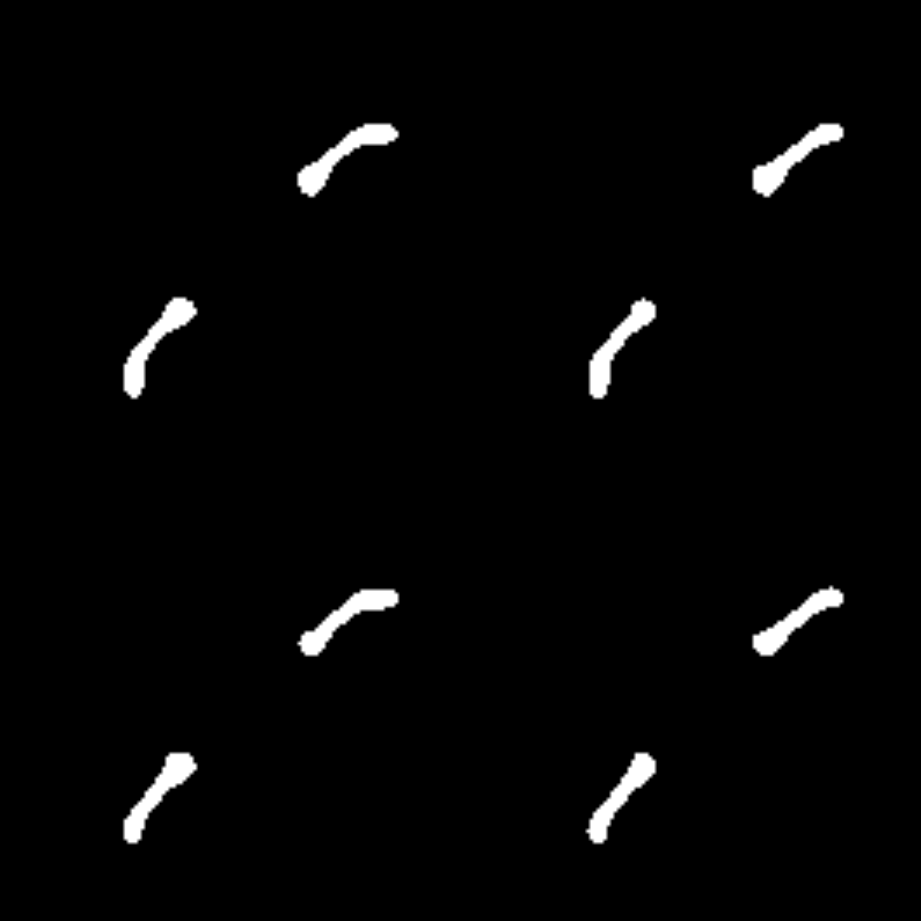}
\caption{}
\label{fig:R1}
\end{subfigure}
\caption{
True positives corresponding to ground truth,~\cref{fig:gt}.
We consider distinct types of loop closures events and how they are and should be interpreted and understood in radar place recognition -- \emph{teach-and-repeat} is shown in \subref{fig:R2}, \emph{revisiting} is shown in \subref{fig:R1}.
}
\label{fig:Rs}
\vspace{-.5cm}
\end{figure}

\cref{fig:Rs} shows decomposition of loop closure events by proposed types.
We argue for the importance of a qualitative distinction of performance for matches arising from:
\begin{enumerate}
\item \emph{teach-and-repeat}, where the taught path is closely followed (in the case of the \textit{Oxford} and \textit{MulRan} datasets where the vehicle is under manual control there may be minor intra-lane variation), and
\item \emph{revisiting}, where the same place is visited on more than one occasion and at arbitrary attitude (in this example, backwards) during a foray.
\end{enumerate}
Specifically, the results presented in~\cref{sec:results} \emph{mask out} from the embedding distance matrix the true positives of \emph{all other} decomposition categories.
We refer to the decompositions as \texttt{rpt} and \texttt{rev}.
Therefore, to isolate e.g. \texttt{rev} in~\cref{fig:Decomp}, we use~\cref{fig:R2} (\texttt{rpt}) to ignore ground truth and predicted matches in~\cref{fig:gt,fig:vTR2_dist}.

\pgfplotsset{
select row/.style={
x filter/.code={\ifnum\coordindex=#1\else\def\pgfmathresult{}\fi}
}
}

\pgfplotstableread[col sep=comma]{
strategy,auc
vT,0.18036769316128445
vR,0.07241790706822228
vTR,0.1847223149569314
vTR2,0.24083457889134752
}\aucMulRanData

\pgfplotstableread[col sep=comma]{
fscore,vT,vR,vTR,vTR2
F0.5,0.24755420349688823,0.08905897180457338,0.2253822955836783,0.3185744964585004
F1,0.23814210140297581,0.13038488560252898,0.24322967607654392,0.304215127338117
F2,0.2599882270008784,0.18754193247134957,0.27846942531116875,0.2931727926835624
}\fScoreMulRanData

\begin{figure}[!h]
\centering

\begin{subfigure}{0.48\textwidth}
\begin{tikzpicture}
\begin{axis}[
ybar,
bar width=.5cm,
width=\columnwidth,
height=0.375\textwidth,
enlarge x limits=0.25,
symbolic x coords = {vR,vT,vTR,vTR2},
xtick=data,
xticklabels={\texttt{vR},\texttt{vT},\texttt{vTR},\texttt{vTR2}},
xlabel=Batch strategy,
ylabel=\texttt{AUC},
grid=major,
xmajorgrids=false
]
\addplot table[x=strategy,y=auc]{\aucMulRanData};
\end{axis}
\end{tikzpicture}
\caption{
Aggregate precision-recall area-under-curve (\texttt{AUC}) metrics for the \textit{MulRan} dataset.
}
\label{fig:pr-mulran}
\end{subfigure}
\begin{subfigure}{0.48\textwidth}
\begin{tikzpicture}
\begin{axis}[
ybar,
bar width=.5cm,
width=\columnwidth,
height=.375\textwidth,
enlarge x limits=0.25,
legend style={at={(0.7,1.4)}, anchor=north,legend columns=-1},
symbolic x coords = {F0.5,F1,F2},
xtick=data,
xticklabels={$F_{0.5}$,$F_{1}$,$F_{2}$},
xlabel=Aggregate precision-recall metric,
ylabel=\texttt{Score},
grid=major,
xmajorgrids=false
]
\addplot table[x=fscore,y=vR]{\fScoreMulRanData};
\addplot table[x=fscore,y=vT]{\fScoreMulRanData};
\addplot table[x=fscore,y=vTR]{\fScoreMulRanData};
\addplot table[x=fscore,y=vTR2]{\fScoreMulRanData};
\legend{\texttt{vR},\texttt{vT},\texttt{vTR},\texttt{vTR2}}
\end{axis}
\end{tikzpicture}
\caption{
F-Score metrics for the \textit{MulRan} dataset.
}
\label{fig:f-mulran}
\end{subfigure}

\caption{
Evaluation of aggregates of precision-recall as well as \texttt{Recall@N} performance for batch strategies over \textit{MulRan}.
}
\label{fig:Datasets}
\vspace{-.55cm}
\end{figure}
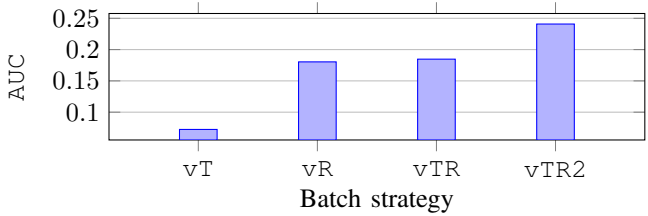
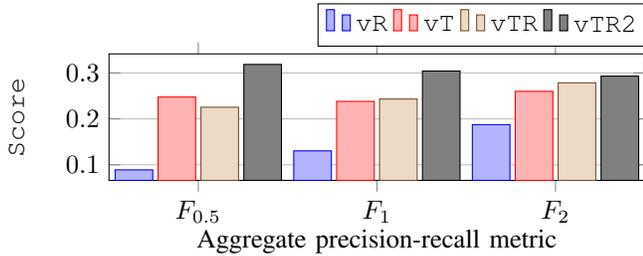

\section{Results}%
\label{sec:results}

\cref{fig:pr-oxf} shows the \texttt{Recall@P} metrics for the \textit{Oxford} dataset.
We achieve from \SI{11.72}{\percent} up to as much as \SI{16.34}{\percent} recall for precisions of \SIrange{99}{95}{\percent}.
This is compared to \SIrange{4.26}{7.08}{\percent} for the most na\"{i}ve constrastive learning approach, \texttt{vR}.
This corresponds to $F_1$, $F_2$, and $F_{0.5}$ scores of \num{0.60}, \num{0.55}, and \num{0.63}, respectively.
Here, we have outperformed the supervised embedding learning of S{\u{a}}ftescu \textit{et al}~\cite{saftescu2020kidnapped}, where on this same test split the corresponding scores were reported as \num{0.53}, \num{0.48}, and \num{0.60}.

\cref{fig:rc-oxf} shows the \texttt{Recall@N} metrics for \textit{Oxford}.
Here, we notice when only using a single database candidate (\texttt{Recall@N=1}), we correctly localise \SI{98.38}{\percent} of the time (\texttt{vTR2}) as compared to \SI{73.13}{\percent} for the baseline, \texttt{vR}.
This is also superior to the results reported in Barnes and Posner~\cite{barnes2020under}, approximately \SI{97}{\percent}.

Note the fidelity of the embedding space for \texttt{vTR2},~\cref{fig:vTR2_dist}, as compared to the ground truth matrix,~\cref{fig:gt}.
Differently to \cref{fig:pr-oxf,fig:rc-oxf}, \cref{fig:bsnn} shows the true and false positives for \texttt{Recall@P=80\%}.
\cref{fig:rnnn} shows such for \texttt{Recall@N=25}, a similar number of database candidates used as in the work of Kim \textit{et al}~\cite{kim2020mulran} as well as Barnes and Posner~\cite{barnes2020under}.
In both, we consider it reasonable to expect a downstream process to disambiguate \emph{some} false matches in order to retain various types of loop closure. 
We see that rotational invariance is well understood by the network trained as we propose.

\pgfplotsset{
select row/.style={
x filter/.code={\ifnum\coordindex=#1\else\def\pgfmathresult{}\fi}
}
}

\pgfplotstableread[col sep=comma]{
N,ra0,ra1
1,98.38383838383838,95.95959595959596
2,99.19191919191918,97.77777777777777
3,99.19191919191918,98.7878787878788
4,99.19191919191918,98.7878787878788
5,99.19191919191918,99.19191919191918
6,99.19191919191918,99.39393939393939
7,99.39393939393939,99.5959595959596
8,99.39393939393939,99.5959595959596
9,99.5959595959596,99.79797979797979
10,99.5959595959596,99.79797979797979
11,99.79797979797979,99.79797979797979
12,99.79797979797979,99.79797979797979
13,99.79797979797979,99.79797979797979
14,99.79797979797979,99.79797979797979
15,99.79797979797979,99.79797979797979
16,100.0,99.79797979797979
17,100.0,99.79797979797979
18,100.0,99.79797979797979
19,100.0,99.79797979797979
20,100.0,99.79797979797979
21,100.0,100.0
}\netArchRCData

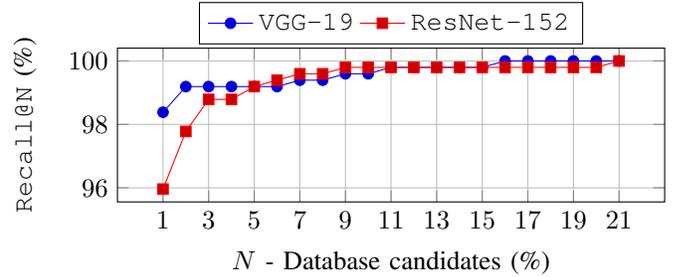
\begin{figure}[!h]
\vspace{-.2cm}
\centering

\begin{tikzpicture}
\begin{axis}[
width=\columnwidth,
height=.2\textwidth,
legend style={at={(0.5,1.3)}, anchor=north,legend columns=-1},
xtick={1,3,5,...,21},
xlabel={$N$ - Database candidates (\%)},
ylabel=\texttt{Recall@N} (\%),
grid=major
]
\addplot table[x=N,y=ra0]{\netArchRCData};
\addplot table[x=N,y=ra1]{\netArchRCData};
\legend{\texttt{VGG-19},\texttt{ResNet-152}}
\end{axis}
\end{tikzpicture}
\caption{
Evaluation of \texttt{Recall@N} for the \texttt{VGG-19} and \texttt{ResNet} architectures over the \textit{Oxford} dataset.
}
\label{fig:netArch}
\vspace{-.3cm}
\end{figure}

\cref{fig:pr-mulran,fig:f-mulran} show the corresponding precision-recall aggregates as well as \texttt{Recall@N} for the \textit{MulRan} dataset.
Scores on this dataset are typically lower than for \textit{Oxford}, which we attribute to lane-level differences as well as the quantity of training data available.
Nevertheless, improvements from \texttt{vR} to \texttt{vTR2} are clearly demonstrated on this challenging dataset, validating each step of the batch strategy in our unsupervised approach.
Indeed, we see a similar incremental improvement in $F_{0.5}$, $F_{1}$, and $F_{2}$ from \texttt{vR} to \texttt{vT}, then to \texttt{vTR} and up to \texttt{vTR2}.
We also plot \texttt{AUC} metrics with a similar improvement with strategy of \num{0.07} to \num{0.24}.

\cref{fig:netArch} shows the performance over the \textit{Oxford} dataset for the two feature extraction networks.
Considering \texttt{Recall@1}, we have only \SI{95.96}{\percent} recall for \texttt{ResNet-152} as compared to \SI{98.38}{\percent} for \texttt{VGG-19}.
On this basis alone, and considering that the gains in the region of \texttt{Recall@6} to \texttt{Recall@11} are much more slight, \texttt{VGG-19} is the preferred architecture.

For comparison against the non-learned method of Kim \textit{et al}~\cite{kim2020mulran}, we list the precision-recall metrics.
These are such that our method achieves $F_1$, $F_2$, and $F_{0.5}$ scores of \num{0.54}, \num{0.50}, and \num{0.54}, respectively.
This is compared to corresponding scores of \num{0.41}, \num{0.37}, and \num{0.45} for \texttt{SC}.

\pgfplotsset{
select row/.style={
x filter/.code={\ifnum\coordindex=#1\else\def\pgfmathresult{}\fi}
}
}

\pgfplotstableread[col sep=comma]{
fscore,vTR2,SC
F0.5,0.54,0.41
F1,0.5,0.37
F2,0.54,0.45
}\DecompPRData

\pgfplotstableread[col sep=comma]{
metric,vTR2,SC
rpt,98.38383838383838,92.32323232323232
rev,17.77777777777778,7.474747474747474
}\DecompRCData

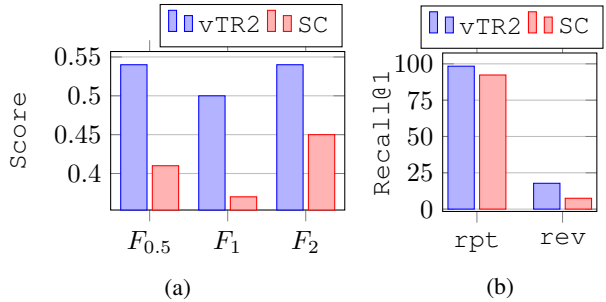
\begin{figure}[!h]
\vspace{-.2cm}
\centering

\begin{subfigure}{0.53\columnwidth}
\begin{tikzpicture}
\begin{axis}[
ybar,
bar width=.35cm,
width=\columnwidth,
height=105pt,
enlarge x limits=0.25,
legend style={at={(0.6,1.3)}, anchor=north,legend columns=-1},
symbolic x coords = {F0.5,F1,F2},
xtick=data,
xticklabels={$F_{0.5}$,$F_{1}$,$F_{2}$},
ylabel=\texttt{Score},
grid=major,
xmajorgrids=false
]
\addplot table[x=fscore,y=vTR2]{\DecompPRData};
\addplot table[x=fscore,y=SC]{\DecompPRData};
\legend{\texttt{vTR2},\texttt{SC}}
\end{axis}
\end{tikzpicture}
\caption{}
\label{fig:DecompPR}
\end{subfigure}
\begin{subfigure}{0.41\columnwidth}
\begin{tikzpicture}
\begin{axis}[
ybar,
bar width=.35cm,
width=\columnwidth,
height=105pt,
enlarge x limits=0.4,
legend style={at={(0.45,1.3)}, anchor=north,legend columns=-1},
symbolic x coords = {rpt,rev},
xtick=data,
xticklabels={\texttt{rpt},\texttt{rev}},
ylabel=\texttt{Recall@1},
ytick={0,25,50,75,100},
y label style={at={(0.2,0.5)}},
grid=major,
xmajorgrids=false
]
\addplot table[x=metric,y=vTR2]{\DecompRCData};
\addplot table[x=metric,y=SC]{\DecompRCData};
\legend{\texttt{vTR2},\texttt{SC}}
\end{axis}
\end{tikzpicture}
\caption{}
\label{fig:DecompRC}
\end{subfigure}
\caption{
Evaluation on \textit{Oxford} of \subref{fig:DecompPR} precision-recall as compared to Kim \textit{et al}~\cite{kim2020mulran} and \subref{fig:DecompRC} \texttt{Recall@1} for decompositions of loop closure event types -- localising along the taught path (\texttt{rpt}) and out-of-lane (\texttt{rev}).
}
\label{fig:Decomp}
\vspace{-.2cm}
\end{figure}

These results are further decomposed as per~\cref{sec:decom} in~\cref{fig:Decomp} and correspond to the \texttt{rpt} matches.
Our method achieves \SI{17.78}{\percent} recall with \texttt{Recall@1} for \texttt{rev} loop closures as compared to \SI{7.47}{\percent} for \texttt{SC}.
We consider that \texttt{SC} is translationally sensitive to out-of-lane driving, rather than being rotationally suspect. 
This shows that our method understands to an extent out-of-lane driving, which we attribute to our learning of rotational invariance and trajectory sampling.
As mentioned in~\cref{sec:motaug} above, out-of-lane robustness could be easily learned in a weakly supervised setting, using GPS/INS as a pseudo groundtruth.
We expect that methods for tackling this in the unsupervised setting could be a focus of future work in this area.

\pgfplotsset{
select row/.style={
x filter/.code={\ifnum\coordindex=#1\else\def\pgfmathresult{}\fi}
}
}

\pgfplotstableread[col sep=comma]{
metric,klvTR2,vTR2
50,78.77576449779981,71.93032552789612
80,44.535590052206345,35.49400342642908
95,17.696381288522424,16.349099215947515
99,17.696381288522424,11.728363013282133
}\prUncertainData

\begin{figure}
\vspace{-.4cm}
\centering

\begin{tikzpicture}
\begin{axis}[
ybar,
bar width=.5cm,
width=\columnwidth,
height=.2\textwidth,
enlarge x limits=0.25,
legend style={at={(0.7,1.3)}, anchor=north,legend columns=-1},
symbolic x coords = {50,80,95,99},
xtick=data,
xlabel={$P$ (\%)},
ylabel=\texttt{Recall@P} (\%),
grid=major,
xmajorgrids=false
]
\addplot table[x=metric,y=vTR2]{\prUncertainData};
\addplot table[x=metric,y=klvTR2]{\prUncertainData};
\legend{\texttt{vTR2},\texttt{KL-vTR2}}
\end{axis}
\end{tikzpicture}

\caption{
Evaluation of precision-recall for embedding families with active dropout and distributional similarity.
}
\label{fig:Uncertain}
\vspace{-.6cm}
\end{figure}
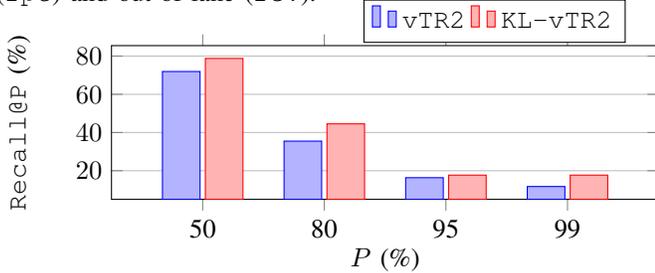

\cref{fig:Uncertain} shows the boosts to performance available from the distributional similarity proposed in~\cref{sec:distsim}.
Here, we use estimated uncertainty to enable or disable putative localisation results.
Using a family of embeddings and distributional similarity boosts our performance to $F_1$, $F_2$, and $F_{0.5}$ scores of \num{0.65}, \num{0.57}, and \num{0.67}, respectively, from \num{0.61}, \num{0.55}, and \num{0.63}.
Correspondingly, we achieve \SI{17.70}{\percent} for \texttt{Recall@P=99\%} as compared to \SI{11.73}{\percent} and \SI{44.54}{\percent} for \texttt{Recall@P=80\%} as compared to \SI{35.49}{\percent}.

\section{Conclusion}%
\label{sec:conclusion}

We have presented the first application of unsupervised learning to radar place recognition.
We bolster the applicability of an invariant instance feature learning approach by sampling temporally proximal scans to act as augmentations of batch instances for the contrastive objective.
These are further augmented by considering the full horizontal field-of-view of the radar sensor and performing random rotations to facilitate out-of-lane loop closures.
Finally, we account for the uncertainty in model predictions and perceptually ambiguous and difficult scenery by measuring distributional similarity between families of radar scan embeddings.
In tests over two sizeable public radar datasets, we outperform three previously published methods which are candidates for the current state-of-the-art, showing the great promise of unsupervised techniques in this area.

\section*{Acknowledgements}

This work was supported by the Assuring Autonomy International Programme, a partnership between Lloyd’s Register Foundation and the University of York as well as EPSRC Programme Grant ``From Sensing to Collaboration'' (EP/V000748/1).
We would also like to thank our partners at Navtech Radar.

\bibliographystyle{IEEEtran}
\bibliography{biblio}

\end{document}